\documentclass[10pt,journal]{IEEEtran}

\usepackage{balance}

\ifCLASSOPTIONcompsoc
  \usepackage[nocompress]{cite}
\else
  \usepackage{cite}
\fi
\ifCLASSINFOpdf
\else
\fi
\usepackage{url}
\newcommand\MYhyperrefoptions{bookmarks=true,bookmarksnumbered=true,
pdfpagemode={UseOutlines},plainpages=false,pdfpagelabels=true,
colorlinks=true,linkcolor={black},citecolor={black},urlcolor={black},
pdftitle={Bare Demo of IEEEtran.cls for Computer Society Journals},
pdfsubject={Typesetting},
pdfauthor={Michael D. Shell},
pdfkeywords={Computer Society, IEEEtran, journal, LaTeX, paper,
             template}}
\ifCLASSINFOpdf
    \usepackage[\MYhyperrefoptions,pdftex]{hyperref}
\else
    \usepackage[\MYhyperrefoptions,breaklinks=true,dvips]{hyperref}
    \usepackage{breakurl}
\fi
\usepackage{graphicx}
\usepackage{wrapfig}
\usepackage{color}
\usepackage{enumitem}
\usepackage{cleveref}
\usepackage{enumitem}
\usepackage{makecell}
\usepackage{xcolor,mdframed}
\usepackage{booktabs}
\usepackage{soul} 

\newcommand{\mytitle}{Real-time face and landmark localization for eyeblink detection}

\begin{document}
%
\title{\mytitle}
%
%
%
%

\author{Paul~Bakker,
        Henk-Jan~Boele,
        Zaid~Al-Ars
        and~Christos Strydis
\IEEEcompsocitemizethanks{\IEEEcompsocthanksitem C. Strydis and H.-J. Boele are with the Department of Neuroscience, Erasmus MC, The Netherlands.\protect\\
Contact e-mail: c.strydis@erasmusmc.nl
\IEEEcompsocthanksitem P. Bakker and Z. Al-Ars are with the Quantum \& Computer Engineering Dept., Delft University of Technology, The Netherlands.}
}

\IEEEtitleabstractindextext{%
\begin{abstract}
    Pavlovian eyeblink conditioning is a powerful experiment used in the field of neuroscience to measure multiple aspects of how we learn in our daily life. To track the movement of the eyelid during an experiment, researchers have traditionally made use of potentiometers or electromyography. More recently, the use of computer vision and image processing alleviated the need for these techniques but currently employed methods require human intervention and are not fast enough to enable real-time processing. In this work, a face- and landmark-detection algorithm have been carefully combined in order to provide fully automated eyelid tracking, and have further been accelerated to make the first crucial step towards online, closed-loop experiments. Such experiments have not been achieved so far and are expected to offer significant insights in the workings of neurological and psychiatric disorders. Based on an extensive literature search, various different algorithms for face detection and landmark detection have been analyzed and evaluated. Two algorithms were identified as most suitable for eyelid detection: the Histogram-of-Oriented-Gradients (HOG) algorithm for face detection and the Ensemble-of-Regression-Trees (ERT) algorithm for landmark detection. These two algorithms have been accelerated on GPU and CPU, achieving speedups of 1,753$\times$ and 11$\times$, respectively. To demonstrate the usefulness of our eyelid-detection algorithm, a research hypothesis was formed and a well-established neuroscientific experiment was employed: eyeblink detection. Our experimental evaluation reveals an overall application runtime of 0.533 ms per frame, which is 1,101$\times$ faster than the sequential implementation and well within the real-time requirements of eyeblink conditioning in humans, i.e. faster than 500 frames per second.

\end{abstract}

}

\maketitle

\IEEEdisplaynontitleabstractindextext

%
\IEEEpeerreviewmaketitle

\section{Introduction}
\label{sec:introduction}

\begin{figure}[t]
    \centering
    \includegraphics[width=\columnwidth]{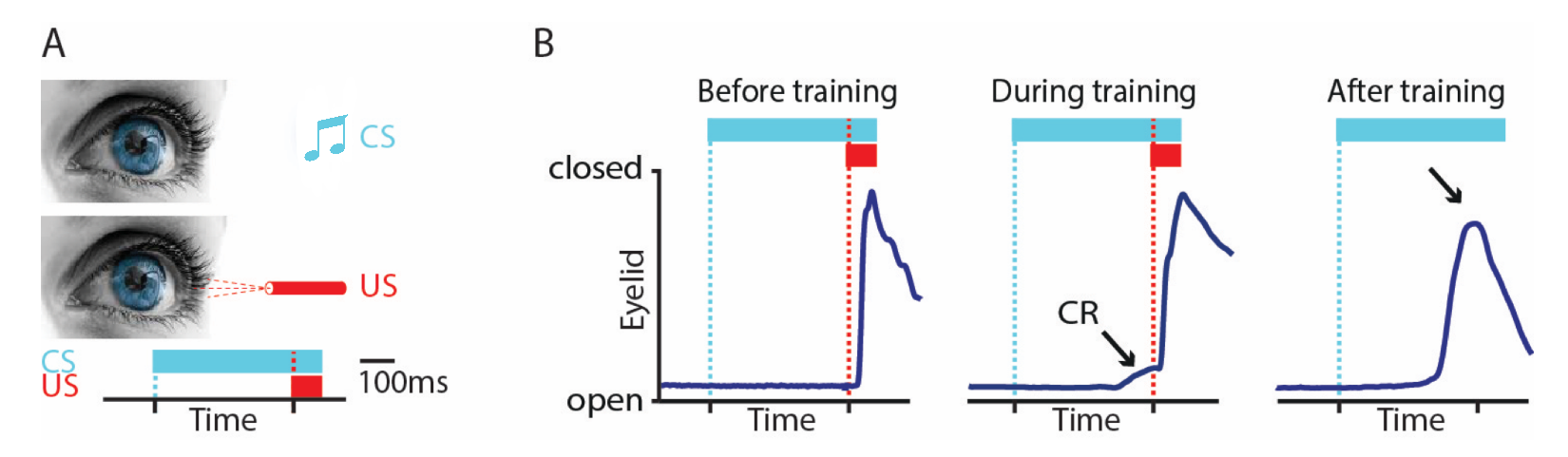}
    \vspace{-0.4cm}
    \caption{Illustration of EBC experiment. The conditioned stimulus (CS) -- panel A -- is a sound paired to an air puff to the eye, the unconditioned stimulus (US). The natural, unconditioned response (UR) to this puff is to close the eyelid. The CS is presented several hundred milliseconds before the US. After a period of training, the naive subject is conditioned to close its eyelid when it hears the sound. The eyelid closure after the sound but before the air puff is the conditioned response (CR); panel B.}
    \label{fig:eyeblinkhuman}
    \vspace{-0.4cm}
\end{figure}

\IEEEPARstart{C}{lassical} conditioning is a learning process that occurs when a behaviorally neutral stimulus is repeatedly paired with a potent stimulus that evokes a particular innate response: the response that was initially elicited by potent stimulus is eventually elicited by the neutral stimulus. One of the best-known examples of this procedure is Pavlov's experiment, in which a dog is conditioned to salivate when hearing the sound of a buzzer~\cite{pawlow1927}. 
Eyeblink conditioning (EBC) is a form of classical conditioning that has been used extensively to study neural structures and mechanisms that underlie learning and memory. The procedure is relatively simple and usually consists of pairing an auditory or visual stimulus (the conditioned stimulus (CS)) with an eyeblink-eliciting unconditioned stimulus (US) (e.g. a mild puff of air to the cornea or a mild shock); see \Cref{fig:eyeblinkhuman}. After many CS-US pairings, an association is formed such that a learned blink, or conditioned response (CR), occurs and precedes US onset. The magnitude of learning is generally gauged by the CR percentage, CR strength, and CR timing.  
Well-trained, healthy individuals can score a high percentage of CRs ($>$90\%), with large amplitudes, and perfect timing, i.e. the eyelid is maximally closed exactly around the onset of the US eye puff. In contrast, conditioning is significantly impaired in humans or animals suffering from neuro-psychiatric disorders like autism, schizophrenia, and ADHD.  
This simple EBC experiment can, thus, provide \textit{a straight-forward, quantitative benchmark of (im)proper brain functionality}.

To capture the eyeblink response (i.e. the ratio of eyelid closure in time), the human subject is recorded with a high-speed camera. Because the subject can move freely, a wider area needs to be recorded and an additional step to identify the region where the eye is situated in each image is needed. At this point, the process of eye localization in each image is slow and requires human intervention.

This work focuses on automating and accelerating the process of eyeblink detection from video in order to achieve real-time processing speeds, at frame rates $\ge 500~fps$. Such a high frame rate is needed so as to reliably detect the onset of the UR blink. This is the first step towards an on-line implementation, which will alleviate the need for off-line video data storage and will allow neuroscientists to manipulate the conditioning experiment interactively, in real-time. Such a feature will have a tremendous impact on the type of neuroscientific questions that can be posed to eyeblink experiments. To the best of our knowledge, this is the first work to ever achieve real-time detection speeds through the combined use of a powerful face- and a landmark-detection algorithm. This paper makes the following contributions:

\begin{enumerate}
	\item Detailed evaluation of three well-known face-detection algorithms w.r.t. accuracy and speed; selection of Histogram-of-Oriented-Gradients (HOG) algorithm.
	\item Literature-based exploration of optimal algorithm for eyelid closure w.r.t. accuracy and speed; selection of the Ensemble-of-Regression-Trees (ERT) algorithm.
    \item Large acceleration and pipelining of HOG and ERT algorithms to achieve first-ever, real-time eyeblink-detection speeds ($\ge 500~fps$).
\end{enumerate}


The baseline and accelerated source code developed for this work is made freely available online\footnote{\url{https://gitlab.com/neurocomputing-lab/eye-blink-conditioning/face-and-landmark-localization-for-eyeblink-detection}.}. This paper is organized as follows: \Cref{sec:background} provides background information on the EBC experiment, object detection in computer vision, and parallel-computing systems. In \Cref{sec:algorithmselection}, different solutions for face detection and eyelid-closure detection are compared and the ones best-suited for this work are selected. \Cref{sec:algorithmpresentation} covers the details of the selected algorithms, while \Cref{sec:implementation} discusses the implementation of their acceleration. In \Cref{sec:evaluation}, the final implementation of the accelerated blink-response detection solution is described and evaluated. Finally, the work is concluded in \Cref{sec:conclusions}.

\section{Background}
\label{sec:background}

\subsection{Human eyeblink conditioning}
\label{subsec:humaneyeblink}

Recording of the eyeblink response has historically been (and still is) done using a potentiometer coupled to the eyelid~\cite{schreurs1997lateralization} or using electromyography (EMG) on the muscle that closes the eyelid~\cite{bracha1997patients}. More recently, the use of computer vision has allowed for a much easier and less invasive way of recording the response.


\begin{figure}[t]
    \centering
    \includegraphics[width=\columnwidth]{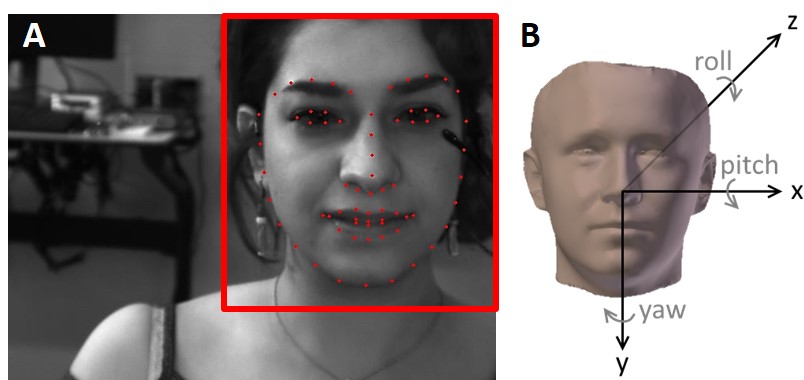}
    \caption{A. Example of desired eyelid detection, identifying the face itself and, then, landmarks on it (including eyelid contours). B. Crucial face-orientation properties for proper detection of eyelid closures.}
    \label{fig:ebcexperiment}
\end{figure}

In the case of the human eyeblink conditioning performed at the Neuroscience department of the Erasmus MC, the subject is facing a camera that is positioned approximately one meter away from the subject. The camera captures the subject's face as well as some surroundings (see \Cref{fig:ebcexperiment}A). This permits the subject some free movement without immediately going out of the camera's scope. The subject's attention is drawn towards the camera (e.g. with a monitor showing a video), to minimize the movement and rotation of the face (see \Cref{fig:ebcexperiment}B). Too much movement or rotation prevents the camera from getting a clear image of the eye and would therefore render the captured video data useless.

For the problem at hand, a combination of object-recognition and -detection algorithms must be used in order to measure the subject as non-invasively as possible, for minimizing disturbance during the conditioning experiment. Furthermore, the selection of algorithms must ensure that the subject has a freedom of movement to a certain degree. Then, by deploying the selected algorithms on a high-performance computing platform, we can achieve the required processing speed in order to perform the detection in real-time. This eschews storing large amounts of data for offline processing, and allows for adjustments during the eyeblink-conditioning experiment depending on the subject's performance.

As will be detailed in Section 3, the combined use of a face-detection and a landscape-localization (specifically for detecting eyelid closures) algorithm has been deemed as the best solution to the problem. The selection of suitable algorithms has been guided by the aforementioned properties of the position and posture of the experiment subject.


\subsection{Real-time eyeblink tracking}
In the course of neuroscientific research, data is often recorded in high speed. This is necessary to capture e.g. the fast movements of mouse whiskers \protect~\cite{mawhiskertracking}, the minute changes in microvasculature blood flow to deduce functional information in awake brains~\cite{koekkoek2018} or, as in this work, eyelid movement. High-performance computing solutions can minimize the time required to process large amounts of data resulting from high-speed (video) recordings, allowing neuroscientists more time to focus on the experiment and, possibly, adjust it in real-time.

As will be discussed in \Cref{sec:implementation}, most of this work focuses on implementing selected algorithms on a Graphics Processing Unit (GPU). This is done by General-Purpose computing on GPU (GPGPU): making use of the many computing cores (many thousand in recent GPU architectures) to compute (parts of) the algorithms in parallel. GPGPU is especially well-suited for leveraging data parallelism, where a single operation is performed on multiple data elements that are independent from each other. They are commonly used to accelerate algorithms arising in life sciences either due to the large computational complexity of the algorithms, such as the case for brain simulations~\cite{michiel_multi_gpu, huang_gpu, brain_frame}, or due to the large amount of data being processed, such as the case for genomics~\cite{ernst16, ernst18, shanshan18}.

\begin{figure}[t]
    \centering
    \includegraphics[width=\columnwidth]{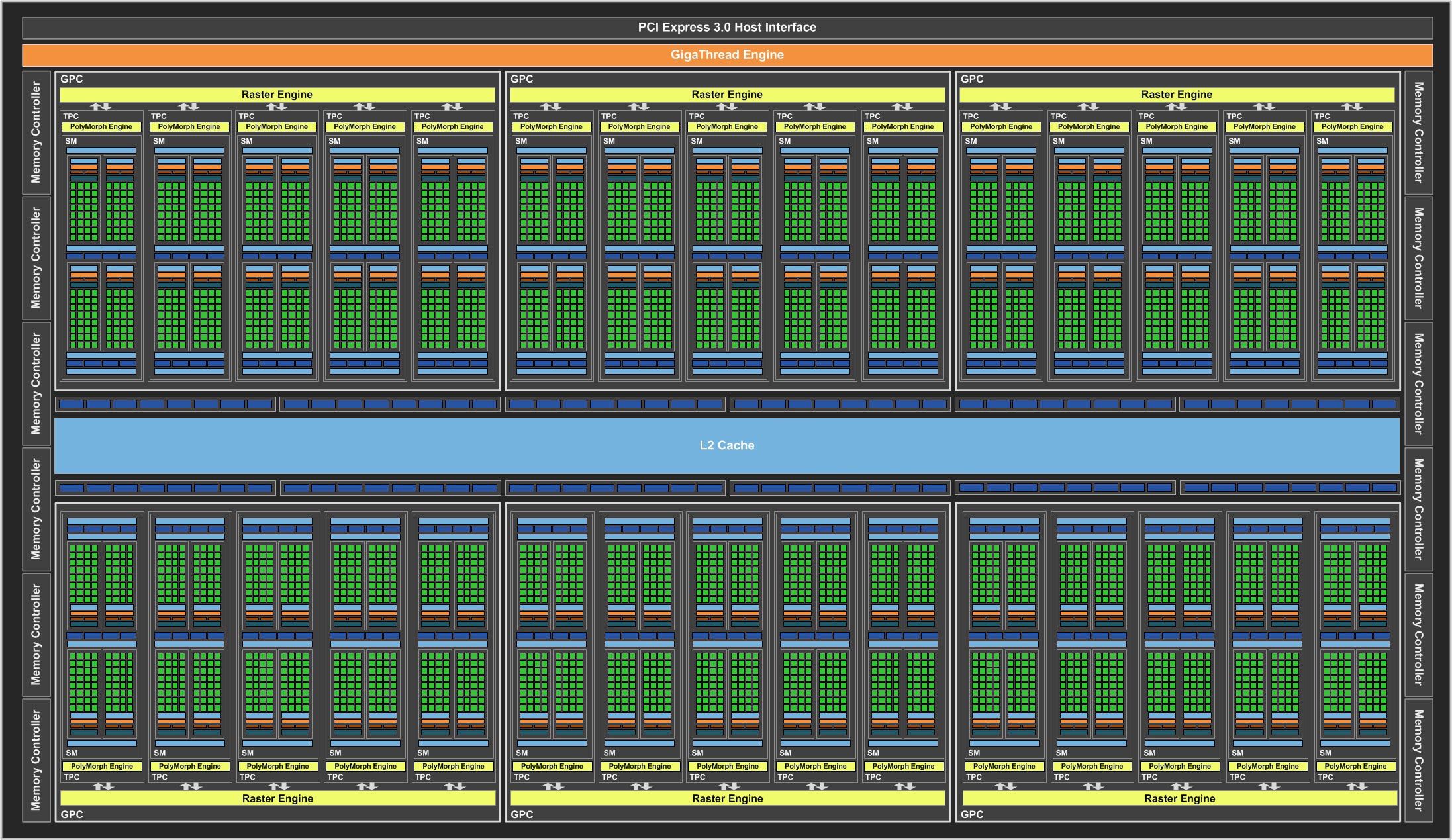}
    \caption{NVIDIA\texttrademark~Pascal GPU architecture. Green dots represent computing cores; 128 computing cores are grouped in one SM.}
    \label{fig:titanxarchitecture}
\end{figure}

The GPU cores are divided among a number of Streaming Multiprocessors (SM), which can execute different tasks independently using \textit{streams}. Streams can be used when not enough data parallelism can be exploited to keep all \textit{cores} occupied with the same task. In this work, each of these SMs consists of 128 computing cores, and can run up to 1024 different \textit{threads}. Threads are executed in groups of 32, called a \textit{warp}. Since each thread has its own set of registers, and there are more threads than cores, context switching is performed to start operations on the next warp while the previous is still waiting to be completed. This is called latency hiding \protect~\cite{volkov2016understanding}. A schematic overview of the GPU architecture used in this work can be seen in \Cref{fig:titanxarchitecture}.

\section{Algorithm selection}
\label{sec:algorithmselection}

In this section, alternatives for both face and eyelid-closure detection shall be investigated. Not only do we need to make sure that the resulting algorithm is always able to detect the eyeblink response in our recording setting, but there is also a significant difference in detection speed between different alternatives. To ensure that the right algorithmic components are selected for our recording environment, a set of requirements was drafted based on hundreds of experiments manually executed within the Neuroscience department of the Erasmus MC. These requirements are summarized in \Cref{tab:facedetectionconditions}.


\begin{table}[t]
    \centering
    \caption{Conditions under which eyeblink-response recording must work.}
    \label{tab:facedetectionconditions}
    \resizebox{0.90\columnwidth}{!}{\begin{tabular}{ccccc}
        \toprule
        \textbf{Yaw}    & \textbf{Roll}    & \textbf{Pitch}    &\textbf{Face size}     & \textbf{Frame rate} \\
        \midrule
        $\pm 20^\circ$    & $\pm 25^\circ$   & $\pm 40^\circ$  & $\ge 20$\% of image         & $\ge500~fps$  \\
        \bottomrule    
    \end{tabular}}
\end{table}

\subsection{Face-detection algorithms}
\label{sec:facedetection}

Face detection has been one of the most studied subjects in computer science and computer vision for over 15 years. The subject has, therefore, a large variety of well-working alternatives to use. Existing techniques can be divided into two categories: so-called rigid templates and deformable-parts models (DPMs)~\cite{facedetectionsurvey}. Rigid templates learn to model the face as a whole, while DPMs model a face by describing it as sum of its parts (nose, eyes, ears, etc.). While DPMs are able to produce excellent results~\cite{mathias2014face,zhu2012face}, they are also known to be very computationally intensive~\cite{zhu2012face, facedetectionsurvey} and take relatively long to train\cite{mathias2014face}. Due to these reasons, DPMs are not deemed suitable for this project and shall therefore not be investigated any further.

Considering the rigid-template models, from a detailed survey~\cite{facedetectionsurvey} three models are chosen for further testing that are very common in face-detection literature and represent, according to the authors of this work, three distinct levels of complexity of face detectors. These are, in order of time of publication and complexity, the Viola-Jones~\cite{violajones}, Histogram of Oriented Gradients (HOG)~\cite{hog} and state-of-the-art Convolutional Neural Network (CNN)~\cite{krizhevsky2012imagenet} face detectors. While the first two make use of pre-determined hand-crafted features to train a classification model, the CNN learns the features to represent the face during the training process.

Since training one of these models is a time-consuming and non-trivial task~\cite{facedetectionsurvey}, and since there are several excellent face-detector implementations available online, in this work we shall refrain from creating our own face detector from scratch. Instead, a face detector will be adjusted to meet the requirements of EBC experiments. Implementations of the open-source Dlib~\cite{dlib09} (HOG, CNN) and OpenCV~\cite{opencv_library} (Viola-Jones) libraries shall be used.   

\subsection{Face-detection-algorithm comparison}
\label{subsec:facedetectioncomparison}

To determine which of the previously discussed algorithms is best-suited for our particular problem, we need to test their accuracy and speed. The three algorithms were tested on the subset of the Annotated Facial Landmarks in the Wild (AFLW) database~\cite{koestinger11a} that meets our EBC requirements regarding the three rotation angles. The ''Wild'' part of AFLW means that these images are taken in real-life situations, where the lighting, position, size and rotation of the faces, background and occlusion are all uncontrolled. The resolution of the images in the database varies, and images can be greyscale or colored.

The images are upscaled to twice their height and width to enable the detection of smaller faces. We evaluate the speed of the three algorithms based on the execution time on a single core of the same CPU, and their accuracy based on their recall and precision percentages:
$$Recall = \frac{TP}{TP + FN}~;~~Precision = \frac{TP}{TP + FP}$$

A true positive ($TP$) is the detection of a face that is annotated in the database, a false negative ($FN$) is when there is a face annotated in the database that is not found by the face detector, and a false positive ($FP$) is when the face detector falsely detects an image region as a face. 


\begin{table}[t]
    \centering
    \caption{Results of face detectors on limited-rotation subset of AFLW database (6079 faces) with images upscaled to double width and height. For comparison, the recall and precision scores on the complete (non-upscaled) AFLW dataset can be seen between parentheses.}
    \label{tab:aflwsubsetupscaled}
        \resizebox{\columnwidth}{!}{\begin{tabular}{lrrrrrr}
        \toprule
        \textbf{Algo.} & \textbf{TP}   & \textbf{FN} & \textbf{FP} & \textbf{Recall (\%)} & \textbf{Precision (\%)} & \textbf{Time (s)} \\ \midrule
        Haar               & 5048          & 1031        & 1339        & 83.0 (41.1)          & 79.0 (84.6)             & \textbf{3172}     \\
        HOG                & 5786          & 293         & \textbf{853}& 95.2 (62.3)          & \textbf{87.2} (90.6)    & 57800             \\
        CNN                & \textbf{5956} & \textbf{123}& 1437        & \textbf{98.0} (87.6) & 80.6 (87.2)             & 241811			\\ \bottomrule	
    \end{tabular}}
\end{table}

\subsubsection{Recall and precision}
\label{subsec:accuracy}
The results of the three face-detection algorithms on the database are concisely presented in \Cref{tab:aflwsubsetupscaled}. 
|Unsurprisingly, CNN achieves the highest score on recall (i.e. sensitivity) as it manages to detect 98.0\% of the faces in the constrained dataset. These models are known to cope well with changes in appearance, pose and illumination as long as the dataset they are trained on contains a large enough amount of varying examples. While the HOG and Haar face detectors show low recall scores on the complete dataset, their scores substantially improve on the limited-rotation subset, with HOG achieving almost the same recall score as CNN. This indicates that these algorithms have more difficulty with object rotations.

The precision scores show a more unexpected picture. Specifically, the precision score for the CNN is much lower than expected. Visual inspection of the falsely detected faces (FP) for each method reveals that the AFLW annotations are indeed not accurate, and miss out on some (mostly smaller, background) faces that are actually in the picture and should be counted as a correctly detected face. To see which of the FPs are in fact valid and which are not, the CNN is used to reclassify the FP images as ``face'' or ``non-face''. With a recall of 98.0\%, the CNN is not perfect, but gives a strong indication of the amount of actual FPs of each method. Both the results of the Haar and HOG face detector can be seen in \Cref{tab:truefalsepos}. Since we cannot use the CNN to reclassify its own FPs, we are not able to re-evaluate its FPs in this way. In the first 96 FPs of the CNN however, every single image depicts a face, and is therefore in fact not a FP but an annotation mistake. This leads us to believe that the precision of the CNN is much higher than the 80.6\% reported in \Cref{tab:aflwsubsetupscaled}, and the highest of the three compared algorithms.

\begin{table}[t]
    \centering
    \caption{Number of FPs re-evaluated with CNN classification because of AFLW annotation errors: FP AFLW indicates the number of FPs according to AFLW-database annotations, while FP CNN indicates the remaining FPs after CNN reclassification has been performed.}
    \label{tab:truefalsepos}
    \resizebox{0.90\columnwidth}{!}{\begin{tabular}{lrrrr} \toprule
        \textbf{Algorithm} & \textbf{FP AFLW} & \textbf{FP CNN} & \textbf{TP} 	& \textbf{Precision (\%)} \\ \midrule
        Haar               & 1344                          & 897                                   & 5048        	& 84.9                    \\
        HOG                & \textbf{791}                  & \textbf{114}                          & \textbf{5786}  & \textbf{98.1}           \\ \bottomrule
    \end{tabular}}
\end{table}

\subsubsection{Speed}
\label{subsec:speed}
Speed-wise, the algorithms show the expected behaviour, as observed speed decreases with the level of algorithmic complexity. The CNN is more than 4$\times$ and 75$\times$ slower than the HOG and Haar algorithms, respectively, while the HOG is more than 18$\times$ slower than the Haar method.

\subsubsection{Verdict on face-detection algorithms}
\label{subsec:conclusionface}
The three face detectors show clear differences in both speed and accuracy. The CNN is the most accurate face detector. It is reliable in varying conditions and settings, and was even able to detect many faces that were not annotated in the AFLW database. However, it is also much slower than the other two methods. In the controlled setting where the eyeblink responses are recorded, the method is somewhat of an overkill. Since we are also focusing on speed, the CNN shall not be be used as face detector in this work. 

Both the Haar and HOG algorithms provide faster alternatives, but where Haar lacks in recall ($3\times$ more false negatives) and precision ($7\times$ more false positives) on the controlled AFLW subset, HOGs accuracy on the AFLW subset is almost on par with the CNN. Although the recall of the HOG algorithm is not perfect, note that the AFLW database contains much harder images than we will typically come across in the controlled eyeblink-recording setting. For the aforementioned reasons, the \textbf{Histogram of Oriented Gradients (HOG)} algorithm shall be used for face detection in this work.  

\subsection{Eyelid-closure detection}
\label{sec:eyedetection}
Once the face has been located in an image, the second step is to detect the eyelid closure. We can approach this problem in two different ways: eye detection followed by eyelid-closure detection or landmark detection where the landmarks on the eyelids are used for eyelid-closure detection.

\subsubsection{Eye detection followed by blink detection}
\label{subsec:eyedetectionfollowedbyblinkdetection}
The first solution requires a similar approach to the face detection. Features are calculated on the part of the image containing the face. A window slides across the face image at different scales, and each subregion is classified as eye or non-eye. Once the image region of the eye has been found, a second algorithm can be used to determine the amount of eyelid closure.

The ideal case would be that the same HOG features used for face detection are also suitable for eye detection. In this way, we would be able to avoid an extra feature-extraction step, as the same features would be used as input for a different classifier. This is not deemed very likely, since different objects often have different features that achieve high classification accuracy.

The extensive survey of~\cite{hansen2010eye} discusses 76 different papers that address eye detection using a video-based approach. Most of the works, however, focus on features that describe the eye in an open state. In fact, the writers conclude that only four of the models show robustness to occlusion due to eye blinks or closed eyes. Furthermore, the survey makes no mention of HOG features used for eye detection, which is a strong indication that we would not be able to re-use the same features that were used for face detection. 

Literature on \textit{blink} detection seems to focus on the detection of an open eye as initialization step, followed by tracking of features or template matching in subsequent frames~\cite{chakka2011competition,polatsek2013eye,tomasi1991detection,morris2002blink,wang2017blink,selvakumar2016real}. While these methods are able to detect the blink of an eye, they do not concern themselves with detecting the amount of eyelid closure. However, once the location of the eye is known, Gabor filters, HOGs or optical-flow features could be used to achieve an estimate of eyelid closure~\cite{minkov2012comparison,boukhers2016shape}. However, this would require an additional step next to the eye-detection step described previously.

\subsubsection{Landmark detection}
\label{subsec:landmarkdetection}
\begin{figure}[t]
    \centering
    \includegraphics[width=\columnwidth]{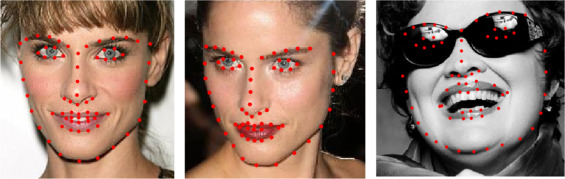}
    \vspace{-0.8cm}
    \caption{Example of landmark detection on three faces~\cite{chen2017recurrent}.} 
    \label{fig:landmarkdetection}
    \vspace{-0.3cm}
\end{figure}

A different approach to the eyelid-closure detection is the use of face-landmark detection, which is also called face alignment. In the survey of~\cite{jin2016face}, 48 different methods of face alignment are discussed and divided into eight classes. Both their accuracy and speed are compared. Most of the methods discussed in the survey focus on the detection of the 68 landmarks shown in \Cref{fig:landmarkdetection}. On each eye, six landmarks are located: two in the corners, two on the upper eyelid and two on the lower eyelid. If these are accurate enough, they could be used directly for the calculation of eyelid closure. In~\cite{soukupova2016real}, these landmarks are already successfully used for blink detection.

The method that seems the most promising is the Ensemble of Regression Trees (ERT)~\cite{kazemi2014one}, as it performs fifth-best in accuracy and second-best in speed, with no other contestant scoring better in both. This method belongs to the class of cascaded regressors, which iteratively refines its estimates of the landmark locations in a number of consecutive stages.

To further review the accuracy of this algorithm, we test its implementation in Dlib on the BioID database~\cite{frischholz2000biold}. This database contains 1,521 images of $384\times 284$ pixels taken in conditions that strongly resemble the ones in our project and of which the location of the center of both eyes is annotated. The algorithm predicts the location of 68 landmarks, and uses these to calculate the center of each eye. Its error is calculated by dividing the euclidean distance between this eye center and the annotated eye center by the inter-ocular distance, and found to be 3.66\%, on average.

\begin{figure*}[t]
    \centering
    \begin{minipage}{.45\linewidth}
        \centering
        \includegraphics[width=0.90\columnwidth]{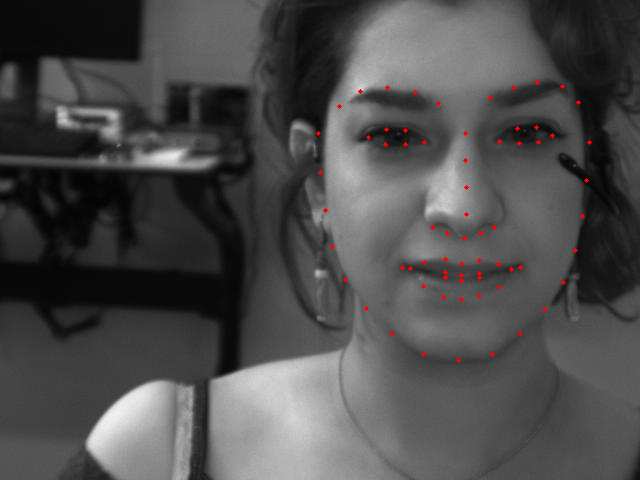}
        \caption{Example of the landmark detection on an image from the ErasmusMC eyeblink-conditioning videos.}
        \label{fig:landmarkerasmus}
    \end{minipage}
    \hspace{.05\linewidth}
    \begin{minipage}{.45\linewidth}
        \centering
        \includegraphics[width=0.95\columnwidth]{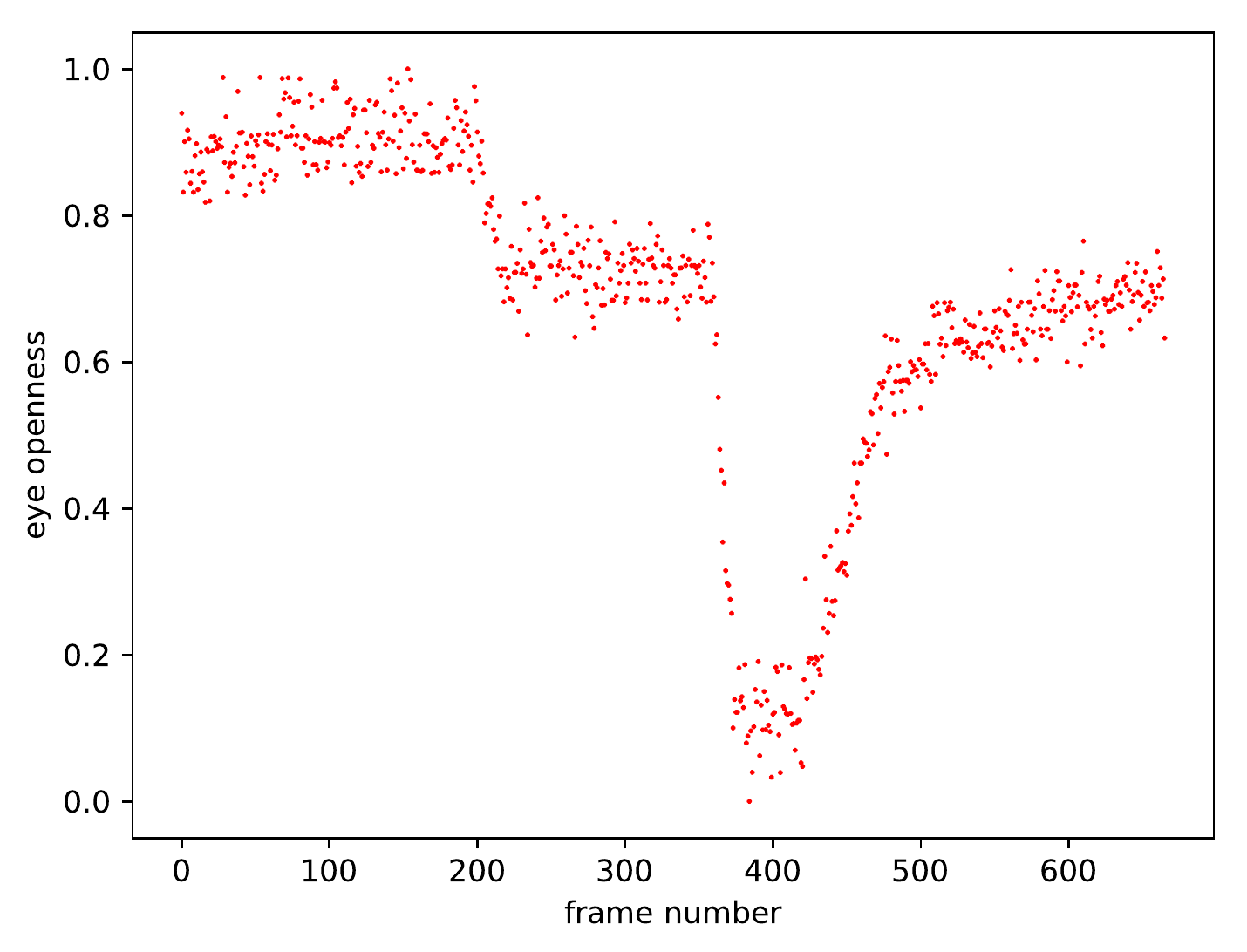}
        \vspace{-0.4cm}
        \caption{Example of an eyeblink-response graph created with the landmark-detection algorithm.}
        \label{fig:landmarkerasmus2}
    \end{minipage}
\end{figure*}

Furthermore, the eyeblink response of multiple blink videos recorded at the ErasmusMC Neuroscience department was plotted using this landmark-detection algorithm. An example of landmark detection on one of these videos, as well as the plotted eyeblink response, can be seen in \Cref{fig:landmarkerasmus} and \Cref{fig:landmarkerasmus2}, respectively. The department neuroscientists were surveyed and deemed the eyeblink-response graphs produced satisfactory. Additionally, the potential for other landmarks to be used to research the behavior of facial muscles during eyeblink conditioning was identified and is a promising future direction.

\subsubsection{Verdict on eyelid-closure-detection algorithms}
\label{subsec:conclusionblink}
Landmark detection and eye detection followed by eyelid-closure detection are two different approaches to detect the eyeblink response. The most promising method under the first approach is a sequence of algorithms that detects the eye, tracks its location and estimates its closure. In contrast, the second approach only requires one algorithm to estimate the landmarks, from which we can directly calculate the level of eyelid closure. In addition, the landmarks on other parts of the face are interesting for research on face-muscle movement during eyeblink conditioning. The ERT algorithm~\cite{kazemi2014one} provides a fast and accurate way of landmark detection and is implemented in the Dlib library. For the aforementioned reasons, the \textbf{Ensemble-of-Regression-Trees (ERT)} algorithm for landmark detection~\cite{kazemi2014one} shall be used for eyelid-closure detection in this work.  

\section{Selected-algorithm details \& profiling}
\label{sec:algorithmpresentation}

Both selected algorithms have been implemented in the Dlib library in C++, and profiling is done on a single core of an AMD Ryzen 7 1800X CPU (3.6 GHz).

\subsection{Face detector}
\label{subsec:facedetectorpresentation}
The HOG face detector of the Dlib library is based on the feature-extraction method of Felzenszwalb et al.~\cite{felzenszwalb2010object} and five classifiers trained on 3000 images of the Labeled Faces in the Wild (LFW) database~\cite{huang2007labeled}. The five classifiers are trained on five different facial rotations to make the face detector more rotation-invariant, and an image is scanned at multiple scales to make the face detection more scale-invariant. In this section, we provide a brief explanation of the HOG algorithm and profile it to see which parts are the most computationally intensive. For a more detailed description, we refer the reader to the work of Felzenszwalb et al.

\subsubsection{Algorithm breakdown}
The original image is scaled down in multiple steps of a $5/6^{th}$-factor until either the width or height of the image is smaller than that of the detection window ($80\times 80$). Note that every step from now on is performed on all scaled versions of the image:

\begin{itemize}
	\item \textbf{Gradient computation}: For every pixel in the image, the gradient orientation is computed and discretized into one of eighteen signed directions.
	
	\item \textbf{Histogramization}: The image is divided into cells of $8\times 8$ pixels. Each cell is described by a histogram of eighteen bins, corresponding to the eighteen discrete gradient orientations of the pixels. Each pixel contributes to gradient-orientation bins of the histograms of the four closest cells. The amount it contributes depends on the gradient magnitude and the distance to the cell.
	
	\item \textbf{Normalization}: The histogram bins of each cell are normalized based on the energy value of the cell and its eight neighbors, where $energy = \sum_{n=0}^{8} (hist\_bin_{n}+hist\_bin_{n+9})^2$.
	
	\item \textbf{Feature computation}: The final feature vector of each cell contains 31 features: 18 signed normalized histogram bins, 9 unsigned normalized histogram bins (where opposite orientations have been added together), and 4 gradient energy features, capturing the cumulative gradient energy of square blocks of surrounding cells. 
	
	\item \textbf{Linear classification filter and detection}: The classifier is trained with a face size of $80\times 80$ pixels, or $10\times 10$ cells. This means that the features of an area of $10\times 10$ cells, 3100 in total, are used as input for the classifier. Each of these features is multiplied by a certain weight, and when the sum of these multiplied features exceeds a threshold, the area is classified as a face. This is done for every area of $10\times 10$ cells in the feature image. As previously stated, there are five classifiers, one for every rotation of the face, so this process is repeated five times.
	
	\item \textbf{Non-maximum suppression}: A final step is performed to reduce a number of detections arising from the same face down to the best-scoring one.
\end{itemize}


\subsubsection{Algorithm profiling \& acceleration} 
\label{subsubsec:facedetalgorithmaccel}
Profiling with Valgrind~\cite{weidendorfer2008sequential} on 10 images from Erasmus MC eyeblink videos ($640 \times 480$ resolution), we see that 33.57\% of the execution time is spent in a function that computes the gradient computation and histogramization steps, while 63.09\% of the time is spent in a function that multiplies the features with the linear classification filter weights. Only functions that contribute more than 10\% of the total execution time are reported, and the average execution time is $0.603~s$ per image.

\begin{figure*}[t]
    \centering
    \includegraphics[width=\textwidth]{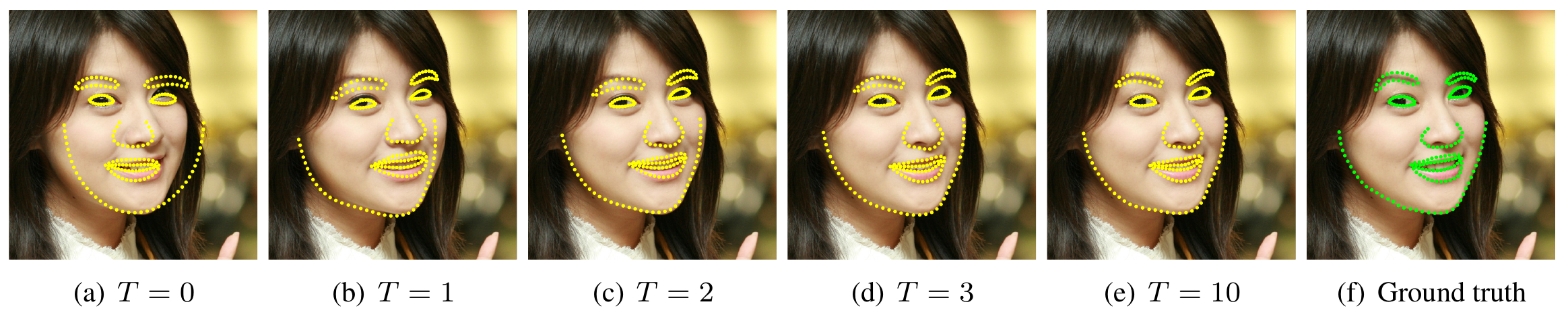}
    \vspace{-0.8cm}
    \caption{Landmark estimates as the number of iterations (T) progresses~\cite{kazemi2014one}.}
    \label{fig:landmarkiterations}
\end{figure*}

The high level of available data-level parallelism of the algorithm (computations on independent pixels, cells and features), combined with the possibility of task-level parallelism across multiple images in a video using GPU streams, seems well-suited for GPU acceleration. To achieve this, we will make use of CUDA~\cite{nickolls2008scalable}.

\subsection{Landmark detector}
\label{subsec:landmarkdetectorpresentation}
The ERT landmark-detection algorithm of the Dlib library is based on the algorithm described by Kazemi and Sullivan~\cite{kazemi2014one} and trained on the iBUG 300-W face-landmark dataset~\cite{sagonas2013300}. 

In contrast to the face-detection algorithm, the landmark-detection algorithm is an iterative process. It is also called a cascaded-regression approach. Each iteration, or level of the cascade, refines the estimate of the landmark locations, as can be seen in \Cref{fig:landmarkiterations} over a number of iterations. For a detailed description of the algorithm, we refer the reader to the work of Kazemi and Sullivan~\cite{kazemi2014one}.

\subsubsection{Algorithm breakdown}
The algorithm starts with an initial estimate of landmark locations that is based on the mean shape of all the images it has been trained with, centered at the middle of the face image. In each of the 15 levels of the cascade, a total of 500 regression trees each calculates a shift of these landmarks to make the detection more accurate. Each regression tree is traversed using the difference between two pixel values at every split, but the critical point of this approach is that these pixel locations are indexed relative to the landmark-shape estimation of the current cascade level. The results of all the regression trees in a level are added to the current landmark estimation, which results in the landmark-shape estimate for the next level. 

This process can be summarized in the following steps:

\begin{itemize}
	\item  \textbf{Initialization}: Initialize the landmark-shape estimate. This is the mean of all landmark shapes with which the detector has been trained.
	
	\item  \textbf{Feature computation}: Calculate the similarity transform between the current shape estimate and the original mean-shape estimate. Use this transformation to calculate the new location of the feature points for the regression trees.
	
	\item  \textbf{Regression-tree estimation}: Traverse each of the 500 regression trees based on pixel intensity differences and add their results to the current landmark-shape estimate.
	
 	\item  \textbf{Repeat}: Repeat the feature computation and regression-tree estimation step for each level of the cascade. 
\end{itemize}


\subsubsection{Algorithm profiling \& acceleration} 
The algorithm is profiled on ten face images extracted by the face detector to identify the computationally intensive steps. The feature computation step (36.79\%), traversing the regression trees (6.32\%) and adding the landmark-shift results of every regression tree to the current landmark-shape estimate (55.68\%) take up the majority of the total execution time of $2.9~ms$ per image.

For the landmark-detection, the data-level parallelism model is less well-suited because of the small amount of regression trees (500) and many stages of the algorithm that would require some form of atomic additions or synchronization. Furthermore, the landmark-detection algorithm is already fast on a single-core CPU, which would make the data transfers to and from an external hardware accelerator relatively costly compared to any potential algorithm speedup.    

Instead, we shall make use of a coarser task-level parallelism approach on a multi-core CPU using OpenMP~\cite{dagum1998openmp}, where every processing element can execute the landmark detection on its own face image without any required communication with the other processing elements. Furthermore, detecting the landmarks on the CPU will allow for overlap with the face detection on the GPU when using pipelining; as will be shown next.

\section{Implementation}
\label{sec:implementation}

The face-detection and landmark-detection algorithms were accelerated using CUDA and OpenMP, respectively, to result in a combined execution time smaller than the $2~ms$ required for this project, due to the $\ge 500-fps$ constraint.

\subsection{Face detection on GPU}
\label{subsec:facegpu}

The face-detection algorithm is optimized for an NVIDIA Titan X GPU, which is powered by the NVIDIA Pascal architecture and has CUDA capability 6.1~\cite{cuda}.

Kernel performance was analyzed and optimized using the NVIDIA Visual Profiler (NVVP). The reported kernel times are the average of 3 runs on an image of $640 \times 480$ pixels.

The most crucial aspect of maximizing GPU performance is memory optimization~\cite{cudabestpractice}. This concerns both the data transfers between the host (CPU) and device (GPU), as well as the memory transactions within the device itself. Since we implemented the algorithm in such a way that data between host and device is only transferred once (each way) per image, the optimizations mostly concern the device memory. These can be summarized in the following steps: 

\begin{itemize}
	\item \textbf{Avoid:} Avoid unnecessary transactions to and from the off-chip memory by making optimal use of the available caches.    
	\item \textbf{Combine:} Combine the necessary memory transactions by coalescing memory requests of the threads within a warp (group of 32 threads).
	\item \textbf{Hide:} Hide the high-latency memory transactions by making use of thread- and instruction-level parallelism. 
\end{itemize}

The implemented face-detection kernels and the optimizations employed are presented next.

\subsubsection{Image-scaling kernel}
The first step of the face-detection algorithm is the image downscaling. The image size of $640 \times 480$ pixels requires a total of 12 scales to detect faces of all sizes, so the image is downscaled 11 times. Each thread in the grid computes the value of a pixel in the downscaled image by means of bilinear interpolation. We have made use of the free bilinear interpolation that comes with storing the image as texture objects in texture memory. This comes with an overhead in creating the texture objects and binding them to the CUDA arrays. Also, to write to CUDA arrays directly from a kernel, we need to make use of surface writes, which requires the creation of surface objects and binding them to the CUDA arrays. However, whenever we are processing a sequence of images with a constant image size, as is the case for our videos, we can re-use the same CUDA arrays and texture and surface objects, which means we only have to create and destroy them once.

Synchronization must occur after each downscaling step (until minimum image size has been reached), because each scaled image is based on the image of one scale higher. The remaining kernels are computed on all image scales, which are completely independent from each other. Synchronization only has to occur between kernels of the same image scale, not in between different image scales.

\subsubsection{Gradient and histogram kernel}
As the first step in the feature-extraction process, each thread in the 2-dimensional grid computes the gradient orientation and magnitude of a pixel, and contributes to the histograms of the four nearest cells. In order to do this, each thread needs to load the pixel values of its two horizontal en vertical neighbors. This spatial locality benefits from the use of CUDA arrays and texture memory. Since the images are already in the texture memory, it also makes sense to keep them there since the conversion from CUDA array to linear device memory requires an extra device-to-device memory transfer. 

Histogramization suffers from a number of difficulties on the GPU. Firstly, the memory location of the histogram bin to which a thread must write is not known at compile time; it depends on the gradient orientation of the pixel. This causes uncoalesced memory writes. Secondly, multiple pixels might write to the same histogram bins at the same time, which could cause race conditions. To prevent this from happening, we need to make use of atomic operations, which sequentializes writes to the same memory address. 

To reduce the number of writes to global memory we combine the results of the threads in a block in shared memory before writing to global memory~\cite{cudafasthistograms}. Moreover, using this intermediate step allows us to perform the writes to global memory in a coalesced way. 

Unfortunately, in the Pascal architecture, atomic addition in shared memory is only hardware-implemented for 32-bit \textit{integers}, while we are working with floating-point numbers. As a workaround, we multiply the floating-point numbers by a factor of $10^{5}$ (maximum while ensuring there will be no overflow) and convert them into integers before the addition, effectively performing a fixed-point addition with 5-decimal-point precision. After the shared-memory atomic additions, the integers are scaled back and converted back to the floating-point data type before writing them to global memory. 



After the histogramization, synchronization is required in order to make sure the histograms of every cell are complete before computing their energy values. The histogram values are stored in linear device memory in a way that ensures coalesced memory loads in the energy and feature computation kernels.

\subsubsection{Energy kernel}
Each thread in the 2-dimensional grid loads the 18 histogram values of a cell in the image and computes its energy value, where:  $$energy = \sum_{n=0}^{8} (hist\_bin_{n}+hist\_bin_{n+9})^2$$ Energy values are stored in a way that ensures coalesced memory loads in the feature computation kernel.

\subsubsection{Feature kernel}
In this kernel, each thread in the 2-dimensional grid loads the 18 histogram values and the energy value of a cell in the image, as well as the energy value of its 8 neighboring cells. It then computes the 31 features per cell as described in~\cite{felzenszwalb2010object}. After the feature kernel, the feature-extraction process is complete. The image is divided into cells of $8\times8$ pixels, each of which is described by a total of 31 features. This is called the \textit{feature image} and is created for every downscaled version of the original image.

\subsubsection{Classifier kernels}
For the classification kernels, the grid is 3-dimensional: there is a thread for every feature of every image cell. Multiplying the feature values with the classification filter weights happens in two steps: 

\begin{itemize}
    \item First, every thread loads a feature of a cell and that same feature of its 9 horizontal (row) neighbors. These feature values are all multiplied with their own classification filter weights, and the results are summed up and stored in an intermediate stage, called the scratch image. 
    \item Then, for every feature of every cell in the scratch image, the same happens, but than on the vertical (column) neighbors. Threads that compute different features for the same image cell all atomically add their result to the same address in global memory. 
\end{itemize}

This results in the \textit{saliency image}: an image where each cell contains a ``face score'', which is the sum of weighted feature values of its $10\times10$ neighboring cells (3100 in total). 
 
The row and column filter values are the same for every image, regardless of their size or specifics. Therefore, we only have to send them from the CPU to the GPU once, where they can remain unchanged until the end of the process. This calls for the use of constant memory, which makes use of a special read-only constant cache that has the ability to broadcast words, which is efficient when multiple threads of a warp have to access the same filter value simultaneously. 

\begin{figure}[t]
    \centering
    \includegraphics[width=\columnwidth]{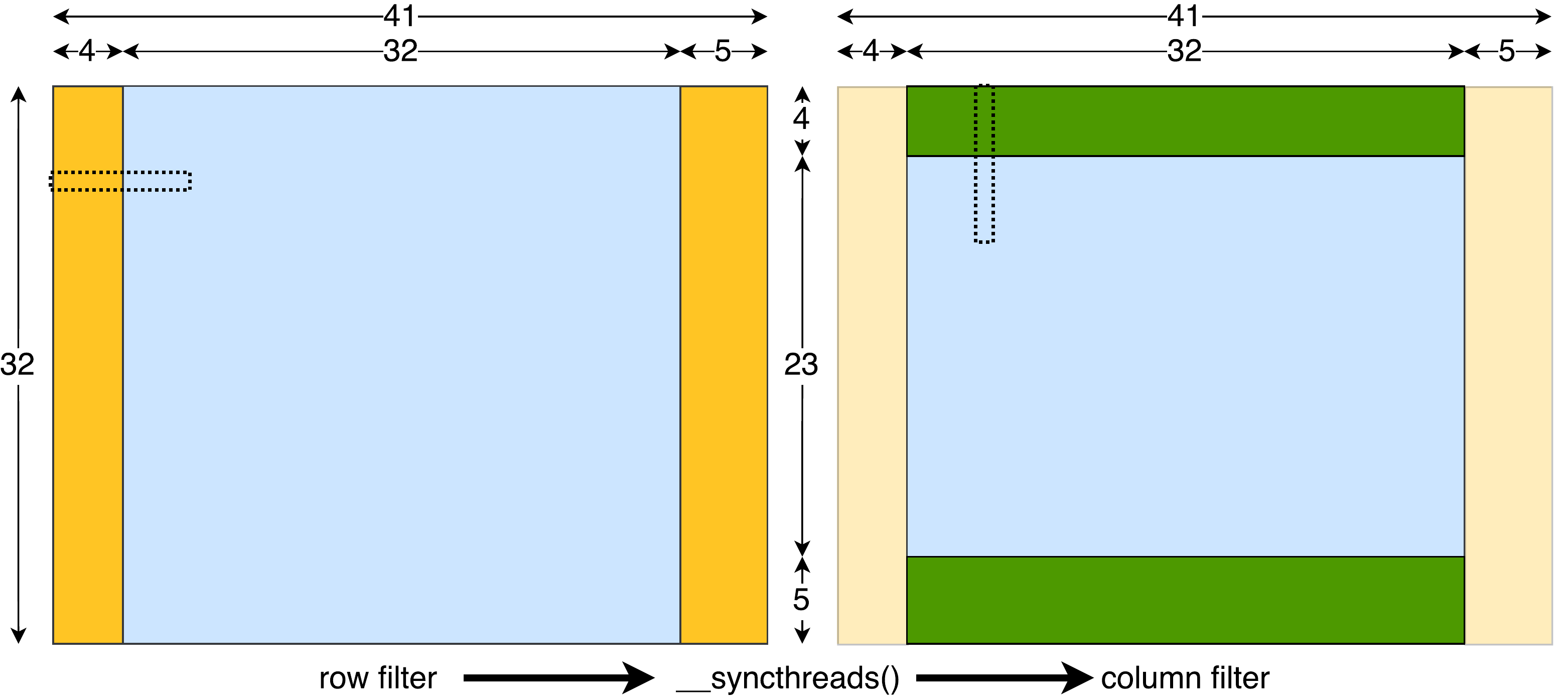}
    \vspace{-0.6cm}
    \caption{Consider a thread block of $32\times32\times1$ threads for the row filter kernel. These threads need an area of $41\times32$ values of a particular feature to calculate the values for the scratch image. Each thread stores its scratch image value in shared memory instead of global memory. Block synchronization with \texttt{\_\_syncthreads()} is performed to ensure that every thread has written its value to shared memory before the row filter computation begins. We can apply the column filter to an area of $32\times23$ without any communication with other thread blocks.}
    \label{fig:rowfiltercombined}
\end{figure}

The synchronization step in between the row and column multiplications requires that the scratch image values calculated by the row filter are stored in, and loaded from, global memory in order to be used by the column filter. This memory-transaction overhead can be avoided by making use of shared memory, block synchronization and by performing redundant computations. This is presented in detail in \Cref{fig:rowfiltercombined}. Synchronizing in this way comes at the cost of redundant computations, as two vertically adjacent thread blocks overlap in an area of $32\times9$ cells in order to compute the column filters for all the cells without inter-block communication. This means that we increase column filter computations by 28.125\% in order to merge the row and column filter kernels, hereby preventing excessive synchronization and global-memory transactions. In addition, the filter values are loaded using \texttt{float2} vector data types which further improves the kernel performance. This process is done for every rotation angle of the head, each with its own filter values, resulting in five different saliency images.

\subsubsection{Detection kernel}
Each thread in a 2-dimensional grid looks at the face score of a single cell. If it is higher than a given threshold, which resides in constant memory, it means a face has been detected, and the thread writes its cell location to global memory. The process happens 5 times, once for every saliency image.

\subsubsection{Non-maximum suppression}
Non-maximum suppression is a technique to reduce multiple detections that arise from the same object to a single highest-scoring one. Since the number of total detections per face is often low (in the order of 1 to 10), this final step does not fit the data-parallel model of the GPU and is performed on the CPU.

\subsubsection{Image sequence with constant size}
\label{subsec:imagesequence}
After all kernels have been optimized, we shift our focus to optimizing the algorithm as a whole for our EBC use case. Regarding memory allocation on the GPU, since we know that we are dealing with a sequence of images that is constant in size, we only have to allocate memory for the images and intermediate steps in the detection process once. This allocated memory can be reused until all images in the sequence have been analyzed. 

Furthermore, all the images in the sequence are $640\times480$ pixels in size, while the required face size is minimally 20\% of the image, as stated in \Cref{sec:algorithmselection}. For the largest image scale, this results in a minimum face size of $128\times96$ pixels. Scaling the image down one time results in a minimum face size of $107\times80$ pixels, which means that we can skip the detection process on the biggest scale of the image without risking to miss faces that should have been detected.

wwFinally, from NVVP output we can see that \texttt{memset}, resetting the arrays in GPU memory to zero, is taking up a large amount of time and interrupting kernel executions. Since we know that these arrays in device memory retain the same size throughout the image sequence, all the arrays in the device memory, except the ones with the original (scaled) images, are combined into one, larger array. A lookup table is kept in device constant memory, which states at which address each of the original smaller arrays begins. This causes all arrays to be reset to 0 with a single \texttt{memset} command, greatly reducing overhead.

\subsubsection{Streams}
\label{subsec:streams} 
Because each image is analyzed on multiple scales, the smaller scales lead to kernels that do not launch enough threads to keep the GPU fully occupied (especially the energy, feature and detection kernels). To prevent SMs from idling, we make use of streams: independent queues of work on the GPU~\cite{cudastreams}. We can place the kernels of each scale in its own GPU stream, so that the execution of multiple kernels from different image scales can be combined when a single one does not make full use of the GPU. 

At first, the improvement over the single-stream implementation was not as much as was expected. NVVP output shows that the concurrent execution of multiple kernels only happens sparsely, even though there are plenty of smaller kernels to be combined. The reason for this is that the CPU is not able to add work to the streams fast enough, in order to build up the available work for when GPU resources are available. Launching a kernel takes roughly 5 to 10 $\mu$s, while some of the smaller kernels have an execution time as little as 1 or 2 $\mu$s. Therefore, there is no `available work' built up for when kernels could be combined, and the GPU is actively waiting for tasks to be added to its streams that it can execute.

\subsubsection{Image combinations}
To tackle this problem, the \texttt{-default-stream per-thread} compiler option was used to employ multiple CPU threads to launch kernels. However, the more threads that were used, the slower the kernel launching time of each individual thread became. Therefore, instead of trying to increase the kernel-launch throughput, the execution time of each kernel was increased by combining images. If the ratio of kernel-launch time to kernel-execution time becomes smaller, the amount of available work on the GPU can build up and kernels can be executed concurrently if resources are available. There are two ways to achieve this:

\begin{itemize}
    \item Increase the amount of work per thread while maintaining the same number of threads.
    \item Increase the amount of threads while maintaining the amount of work per thread.
\end{itemize}

To achieve this, the algorithm had to be adjusted so that each kernel performs its computations on a number of images. The first approach is to loop over a number of images (stored in device memory) from within the kernel, increasing the amount of work per thread. This, however, caused a higher register pressure, lowering the achieved occupancy for some kernels. 

A second approach is to combine multiple images into one, larger image containing multiple faces. The GPU performs face detection, then, on this combined image as if it is any other image and returns an array of detections, and based on their location in the combined image, it can be tracked to which of the original images they belong. Although this increases the number of threads per kernel launch, it also causes an unacceptable loss of detection accuracy, since pixels of a neighboring image can reduce the classification score of a face near a border. 

\begin{figure}[t]
    \centering
    \includegraphics[width=0.90\columnwidth]{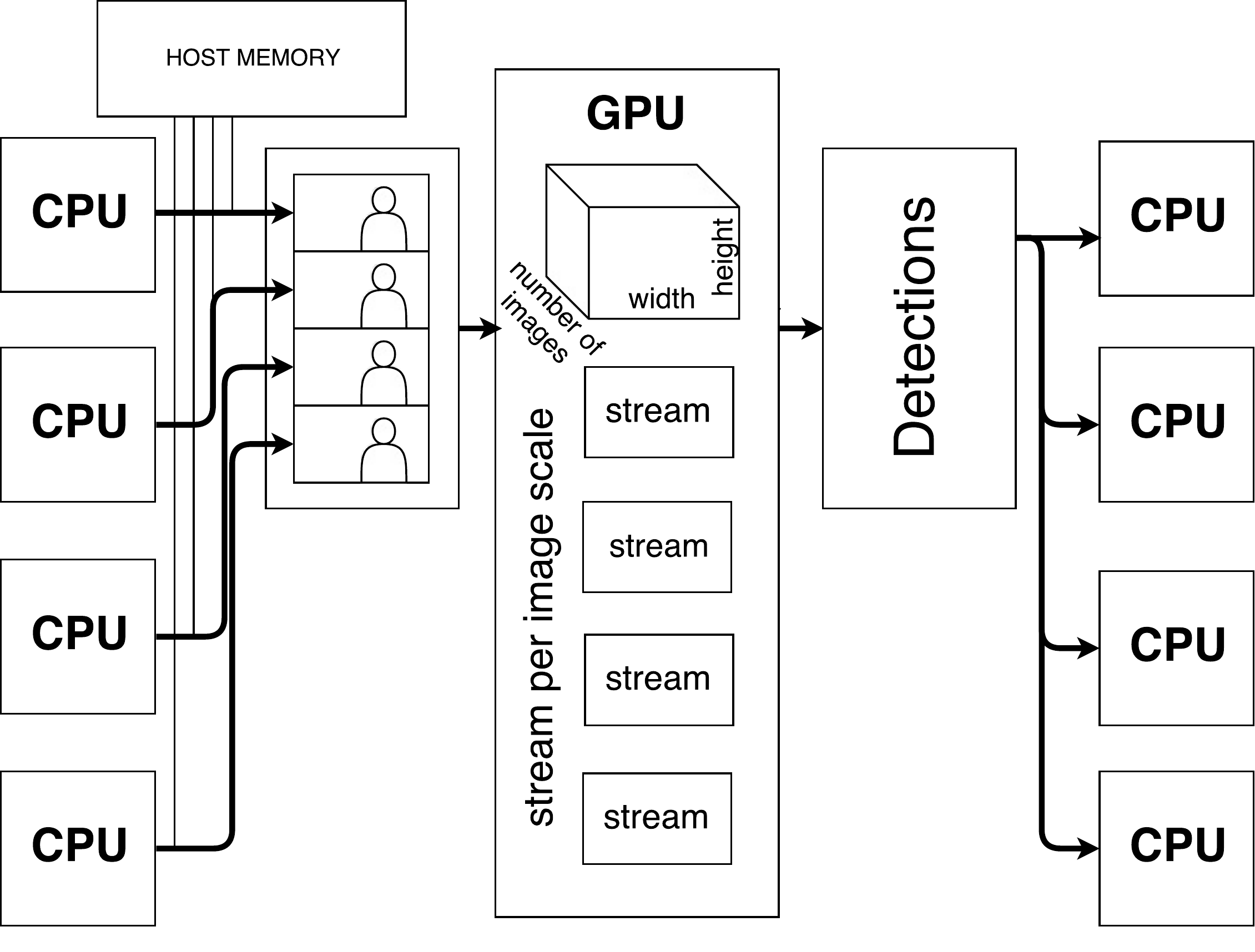}
    \caption{Face-detection implementation where the grid of each kernel increases in the z-dimension, for multiple images analyzed per kernel launch.}
    \label{fig:imagecombinedimension}
\end{figure}

The final approach is a slight variation on the \texttt{for-loop} approach: instead of looping over multiple images inside the kernel, the number of dimensions of the thread grid can be increased by one. The new dimension tells each thread on which image it should perform its computations. This causes an increase in threads per kernel launch, therefore increasing the kernel execution time, while maintaining detection accuracy. The implementation is visualized in \Cref{fig:imagecombinedimension}. The NVVP output confirms that this leads to concurrent kernel execution. Furthermore, performing computations on \textit{16} images per kernel launch leads to the best results. Each of these images gets loaded from host memory by a different CPU thread. 

\subsubsection{Block sizes}
\label{subsec:blocksizes}
The best-performing block sizes are determined empirically for each kernel and shown in \Cref{tab:blocksizeszz}. The $z$-dimension of the thread block is kept at 1 since threads in the same block should perform computations on the same image.

\begin{table}[t]
    \centering
    \caption{Best-performing block sizes for each kernel}
    \label{tab:blocksizeszz}
        \resizebox{.55\columnwidth}{!}{\begin{tabular}{lr}
        \toprule
        \textbf{Kernel}   & \textbf{Block size}  \\
        \midrule
        Scaling           & $16\times8\times1$   \\
        Gradient-Histogram & $32\times16\times1$ \\
        Energy            & $32\times32\times1$  \\
        Feature           & $32\times32\times1$  \\
        Classifier        & $16\times32\times1$  \\
        Detection         & $32\times32\times1$  \\
        \bottomrule
        \end{tabular}}
\end{table}

\subsubsection{Data transfers between host and device}
\label{subsec:datatransfers}
Host-to-device data transfers are roughly 5 MB for 16 images, while the array that is returned from the device to the host (which contains the face locations) is approximately 2 MB. Combined, the data transfers between host and device take up 12.9\% of the total face-detection time. Speeding up transfer time and overlapping kernel executions with memory transfers requires the use of page-locked memory, which is more expensive to allocate and deallocate~\cite{pinnendvsnonpinned}. Because we are dealing with relatively small and infrequent data transfers, it is expected that effectively little will be gained from the use of page-locked memory. Therefore, this option shall not be explored further at this point. 

\subsection{Landmark detection on multi-core CPU}
\label{sec:landmarkcpu}
Once the face has been detected, eyeblink detection can start. A total of 68 landmarks is detected on each face, 6 of which are on the eye and are used to estimate the closure of the eyelid. The OpenMP API (OMP) shall be used for multi-core CPU acceleration of the landmark-detection algorithm. Since there is no dependency between different frames of a video, each frame can be processed individually by a separate CPU thread. To achieve this, we make use of OMP work-sharing constructs.

The OMP work-sharing construct makes use of \texttt{\#pragma omp parallel for} to distribute the iterations of a \texttt{for-loop} among the available CPU threads. In our case, the total amount of frames is divided among the CPU threads. The use of 16 CPU threads resulted in the fastest landmark-detection time.

\subsection{Pipelining}
\begin{figure}[t]
    \centering
    \includegraphics[width=0.85\columnwidth]{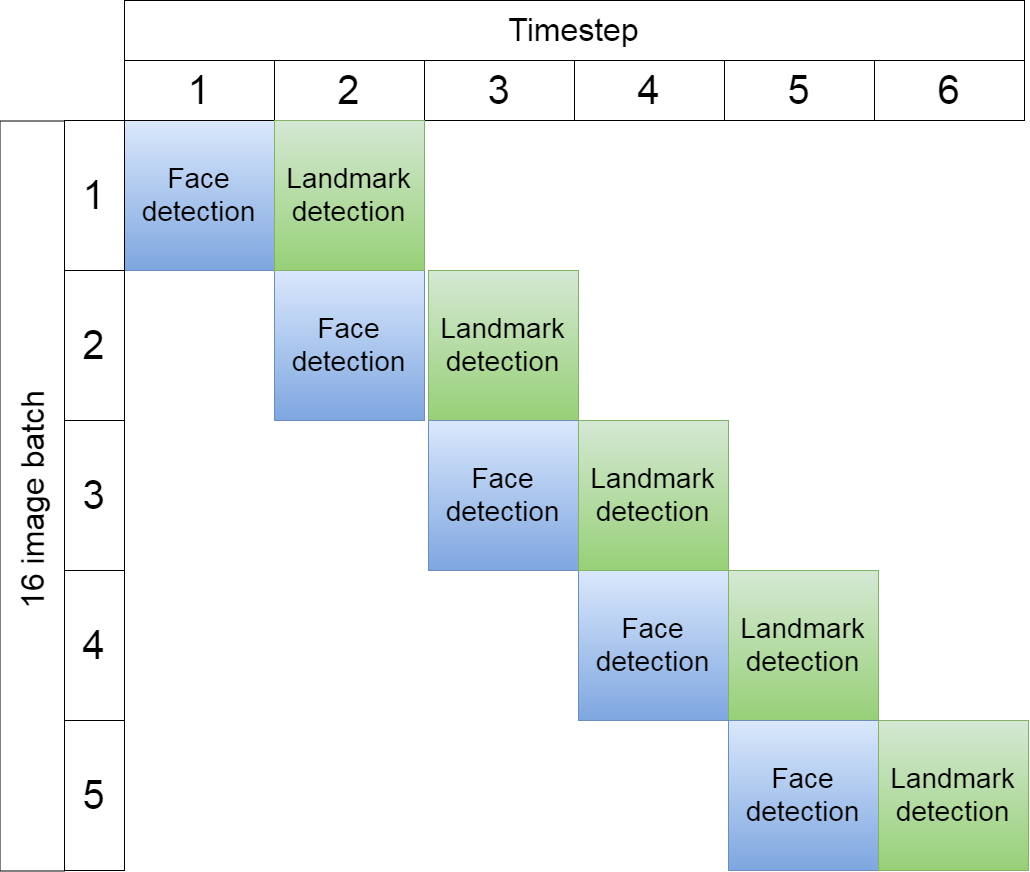}
    \caption{Performing computations on CPU and GPU simultaneously by making use of a pipeline model. Items shown in green are performed on CPU, while items in blue are performed on GPU.}
\label{fig:pipeline}
\end{figure}

Since the face and landmark detections are performed on different computing fabrics, we can (partly) overlap their executions by making use of a pipeline model. This is visualized in \Cref{fig:pipeline}.

\section{Evaluation}
\label{sec:evaluation}

\subsection{Experimental set-up}
\label{subsec:experimentalsetup}
The implementations in this section are evaluated on a set of 10 eyeblink videos, comprising a total of 6,720 grayscale images, recorded at the Neuroscience department of the Erasmus MC. Each image is $640\times480$ pixels in size. The specifications of the CPU and GPU that were used are listed in \Cref{tab:specifications}. Timing measurements on the GPU were performed with NVVP, while on the host were done with the OMP wall-clock timer.

\begin{table}[t]
\centering
\caption{Specifications of device (NVIDIA Titan X (Pascal)) and host (AMD Ryzen 7 1800X + Corsair Vengeance DDR4 DRAM)}
\label{tab:specifications}
    \resizebox{0.75\columnwidth}{!}{\begin{tabular}{lrr}
    \toprule
        \textbf{Specification} & \textbf{Device} 				  & \textbf{Host} 		     	\\
        \midrule
        Clock speed       & 1.53 GHz                         & 3.60 GHz                        \\
        Number of cores        & 3584                             & 8 (16 threads)             \\
        DRAM size              & 12 GB                            & 32 GB                      \\
        DRAM bandwidth         & 480 GB/s                         & 34 GB/s                    \\
        L1 data cache size     & $28\times48$ KB                  & $8\times32$ KB             \\
        L2 cache size          & 3 MB                             & $8\times512$ KB            \\
        L3 cache size          & -                                & 16 MB                      \\
        \bottomrule
    \end{tabular}}
\end{table}

\subsection{Face detection on GPU}
\label{subsec:facedetection}
Three different face-detection implementations are evaluated, and the results are shown in \Cref{tab:facedetfinalresults}. Compared to the original sequential version, a final speedup of 1,753$\times$ is achieved with the optimized, GPU-accelerated version. If we exclude the memory transfers from host to device and back, and purely look at the kernel execution time, the achieved speedup is 2,018$\times$.

\begin{table}[t]
    \centering
    \caption{Performance comparison of different implementations of the face-detection algorithm. Speedup is calculated relative to the slowest implementation.}
    \label{tab:facedetfinalresults}
        \resizebox{\columnwidth}{!}{\begin{tabular}{lrr}
        \toprule
        \textbf{Implementation}                     & \textbf{Time/image (ms)} & \textbf{Speedup} \\
        \midrule
        Original sequential                     & 583.623                      & -            \\
        Unoptimized naive GPU               & 16.624                       & 35           \\
        Optimized GPU incl. mem. transfer & 0.333                        & 1753         \\
        Optimized GPU excl. mem. transfer & 0.289                        & 2018         \\
        \bottomrule             
        \end{tabular}}
\end{table}

The accuracy of the GPU-accelerated face detector in the recording setting of this project was verified on 1,062,080 images with faces from the Erasmus MC eyeblink videos. In only 184 of those images the face detector was unable to find a face in the image, which is less than 0.02\%. In each of these 184 images, the subjects showed a clear violation (too much head rotation) of the face-detection requirements described in \Cref{tab:facedetectionconditions}. On the remaining 1,061,896 images, exactly one face was detected, which indicates the algorithm has a low chance of false positives. 

\subsection{Landmark detection on multi-core CPU}
A task-level parallel approach for landmark detection was implemented with OMP, making use of 16 CPU threads. This implementation is compared to the original sequential implementation in \Cref{tab:landmarkfinalresults}.

\begin{table}[t]
    \centering
    \caption{Performance comparison of original sequential landmark-detection algorithm with the OMP-accelerated version. Speedup is calculated relative to the slowest implementation.}
    \label{tab:landmarkfinalresults}
        \resizebox{0.85\columnwidth}{!}{\begin{tabular}{lrr}
        \toprule
        \textbf{Implementation}   & \textbf{Time/image (ms)}     & \textbf{Speedup}\\
        \midrule
        Original, sequential      & 2.878                        & -                \\
        OMP (16-thread) accelerated & 0.251                        & 11.49            \\
        \bottomrule 
        \end{tabular}}
\end{table}

We achieve a speedup of 11.49$\times$ compared to the sequential implementation by making use of 16 CPU threads. The eyelid-response graphs generated with the landmark locations extracted through the algorithm were reviewed and deemed satisfactory by the neuroscientists of Erasmus MC for which this project has originally been carried out.

The standalone accelerated face-detection (0.289 ms) and landmark-detection (0.251 ms) algorithms achieve similar execution times. Due to the fact that they are decoupled components that are executed on different platforms, pipelining of the tasks of the two components (across CPU and GPU) is possible. The fact that the execution times are well-matched means that the amount of time the CPU or GPU is idle is minimal. 

\subsection{Combined implementation for eyeblink detection}
We compare three versions of the complete eyeblink implementation, containing both the face and landmark detection, in \Cref{tab:eyeblinkfinal}. 

\begin{table}[t]
    \centering
    \caption{Comparison of different implementations of the complete eyeblink-detection solution. Speedup is calculated based on the slowest implementation.}
    \label{tab:eyeblinkfinal}
        \resizebox{0.85\columnwidth}{!}{\begin{tabular}{lrr}
        \toprule
        \textbf{Implementation}    & \textbf{Time/image (ms)}     & \textbf{Speedup} \\
        \midrule
        Original, sequential       & 586.906                      & -                \\
        Accelerated, non-pipelined & 0.801                        & 732              \\
        Accelerated, pipelined     & 0.533                        & 1101             \\
        \bottomrule
    \end{tabular}}
\end{table}

In \Cref{tab:facedetfinalresults}, we can see that the original implementation is heavily dominated by the face-detection component. Acceleration results in comparable face-detection and landmark-detection times, which are overlapped in the final implementation. This results in a final speedup for the complete solution of 1,101$\times$, compared to the original sequential implementation. This enables an eyeblink-detection speed of roughly $1,876~fps$, which more than satisfies the original requirement of $500~fps$. Furthermore, it is noticed that although the face- and landmark-detection algorithms have comparable execution times, overlapping their execution by making use of a pipeline model does not lead to halving the total execution time. This is due to host memory I/O and image-decode operations.  

\subsection{Minimum hardware to meet the requirements}
The current implementation is more than able to achieve the required detection speed of 500 frames per second. Because possible future work for this project could be to investigate a more mobile solution, an estimation is made of the minimum required hardware to achieve a detection speed of $500~fps$.

A detection speed of $500~fps$ equals a maximum detection time of 2 ms. Because of the pipelined implementation, both the steps taken on the CPU (image-decoding and landmark detection) and GPU (face detection) need to be executed in under 2 ms.

The total amount of execution time spent on the CPU for the image-decoding and landmark-detection steps combined -- for a varying number of threads -- is shown in \Cref{tab:cputimes}. From this table we can deduce that, to satisfy the execution time requirement of $<2$ ms, no more than 3 CPU threads need to be utilized. 

\begin{table}[t]
    \centering
    \caption{Execution time of the steps performed on the CPU for a varying number of threads.}
    \label{tab:cputimes}
    \resizebox{.50\columnwidth}{!}{\begin{tabular}{rr}
        \toprule
        \textbf{CPU threads} & \textbf{Time/image (ms)} \\
        \midrule
        1                    & 3.995                        \\
        2                    & 2.395                        \\
        3                    & 1.640                        \\
        4                    & 1.303                        \\
        8                    & 0.761                        \\
        16                   & 0.416                       	\\
        \bottomrule
    \end{tabular}}
\end{table}

The scaling of the execution time on the GPU is somewhat harder to assess, since each of the 6 kernels will scale differently with changes in the GPU architecture. However, NVVP reports that the kernels are either limited by (L2-cache) bandwidth (feature, energy and classifier kernels) or latency (scaling, gradient-histogram and detection kernels), and not by a lack in single-precision floating-point computing performance. This indicates that a reduction in bandwidth would have more severe implications on the face-detection time than a reduction in peak singe-precision floating-point computing performance. Assuming the same interconnect between host and device, the total data-transfer time remains constant at 0.044 ms per image. This leaves 1.956 ms for kernel execution time. This is a factor $\frac{1.956}{0.289}\approx6.77$ more than the kernel execution time of the current implementation. 

\subsection{Problem scalability}
\label{sec:problemscalability}
In the current eyeblink-conditioning setup, videos are recorded at a resolution of $640\times480$ pixels. Because future experiments may be recorded at a higher resolution, it is important to assess the performance with relation to the input image size. 

The different feature-extraction and -classification steps of the HOG algorithm all have linear time complexity $O(n)$, where $n$ is the number of pixels in the image. However, the pyramidal image-scaling structure of the face detector results in a greater number of scales to be analyzed as the input image size increases. While the image size increases, the detection-window size remains $80\times80$ pixels wide, therefore allowing the detection of faces that are relatively smaller compared to the total image size. However, as described in the requirements of \Cref{tab:facedetectionconditions}, the size of the face is always more than 20\% of the image, no matter the input image size. We can use this information to skip the face detection on bigger image scales, which is already done in the current implementation as described in \Cref{subsec:imagesequence}. This effectively results in a feature-extraction and classification time \textit{independent of the input image size}. The only two operations that scale (linearly) with the image size are the host-device memory-transfer time and the scaling-kernel time (which needs to scale the image down more, in order to get to the smaller scales from which the features are extracted and classified). Furthermore, in future work, the algorithm could scale the original image down to the largest image size that needs to be analyzed in one step, instead of multiple steps of $5/6^{th}$ (the scaling-factor, see \Cref{subsec:facedetectorpresentation}), before proceeding with the downscaling in the smaller steps for all image scales that need to be analyzed. This would result in constant time complexity for the image downscaling step.

The ERT algorithm has a runtime complexity that does not depend on the input size but on the number of layers in the cascade (T), the number of regression trees (K) in each layer and the depth of each tree (F); thus, $O(TKF)$, as described in~\cite{kazemi2014one}.

The results for image sizes of $640\times480$, $1280\times960$ and $2560\times1920$ pixels are shown in \Cref{tab:biggerimage}. As the table reveals, the image size increases by a factor 4 per column. The face- and landmark-detection steps show the expected scaling behavior. In case the data-transfer time between host and device becomes the dominant factor of the face detection, it is worth re-evaluating the use of pinned memory for increased transfer speed and concurrent data transfers with kernel executions.

\begin{table}[t]
    \centering
    \caption{Execution time of different algorithmic steps of the blink-response detection for different image sizes.}
    \label{tab:biggerimage}
    \resizebox{0.90\columnwidth}{!}{\begin{tabular}{lrrr}
        \toprule
        \textbf{Algorithmic step} & \multicolumn{3}{c}{\textbf{Time/image (ms)}}                               \\ \midrule
         									   & \textbf{640$\times$480} & \textbf{1280$\times$960} & \textbf{2560$\times$1920} \\ \midrule
        Face detection (HOG)                   & 0.333                   & 0.734                    & 1.436                     \\
        Landmark detection (ERT)               & 0.251                   & 0.268                    & 0.300                    	\\ \bottomrule
    \end{tabular}}
\end{table}

\subsection{Discussion}
\label{sec:discussion}
While the accuracy of the face and landmark detection could be verified with the use of annotated databases, the final result, the \textit{eyeblink-response graph} containing the amount of closure of the eyelid over time, is harder to objectively assess. The results were manually inspected and verified by expert neuroscientists of Neuroscience department at the Erasmus MC, yet the amount of samples that could be evaluated in this way was limited. In any case, landmark detection has been confirmed to accurately determine the location of the eye. If it turns out in the future that the landmarks on the eyelid are eventually not deemed accurate enough for the eyeblink-response graph, an extra algorithm can be applied on the image of the eye to estimate the amount of eyelid closure in a different way.

\section{Conclusions}
\label{sec:conclusions}

\subsection{Contributions}
\label{sec:contributions}

The goal of this work has been to select a combination of algorithms that is able to detect the amount of human-left-eyelid closure in video data (eyeblink-response detection), and to accelerate these algorithms in order to achieve real-time processing speeds ($500~fps$). This is the first step towards an on-line implementation that would not only alleviate the need for large off-line data storage, but also enable neuroscientists to dynamically adjust eyeblink-conditioning experiments based on immediately available feedback on the subject's performance. The detection process is split up in two different parts: the first algorithm detects the location of the face in an image, which is then used as the input to a second algorithm that determines the amount of eyelid closure.

Concisely, the following contributions have been made by this work:
\begin{itemize}
	\item Different face detectors have been compared on the Annotated Facial Landmarks in the Wild database of 24384 face images. The Histogram of Oriented Gradients (HOG) was selected as the algorithm that best-suited the project requirements because of its combination of accuracy and speed.
	
	\item The face-detection algorithm was accelerated on a GPU by making use of data- and task-level parallelism, achieving a speedup of 1,753$\times$ and an execution time of 333 $\mu s$.
	
	\item An Ensemble of Regression Trees (ERT) landmark-detection algorithm was selected to estimate the amount of closure of the eyelid.
	
	\item The landmark-detection algorithm was accelerated with OpenMP on a 16-threaded CPU by making use of a task-level parallelism. This resulted in a speedup of 11.49$\times$ and an execution time of 251 $\mu s$.
	
	\item The accelerated face- and landmark-detection algorithms were combined and their execution was overlapped by making use of a pipeline model. This final eyeblink-response detection algorithm is 1,101$\times$ faster than the original sequential version and achieves a detection speed of 1876 FPS, which is faster than the minimum requirement of 500 FPS needed for real-time processing. 
\end{itemize}

\subsection{Future work}
\label{sec:futurework}
One of the next steps is to move from the current, real-time implementation to an on-line implementation. This would require the recording equipment to directly send the video data as batches of images to a worker node over Ethernet. The worker node will process the images and send the results to an additional node that will show and store the results to allow for immediate analysis of the subject's performance.

Another direction that has been discussed is the implementation of the algorithm on a mobile platform. The current implementation achieves a detection speed of $1,876~fps$, and therefore shows that the required detection speed of $500~fps$ can also be achieved with less powerful hardware. However, it is considered unlikely that the implementation in its current form is able to run on a mobile platform. Therefore, it can be investigated whether the accuracy of the implementation can be reduced to an acceptable extent, in order to speed up the eyeblink-response detection. To do so, we must exploit the high correlation between subsequent images in a video. For example, if we limit the amount of scales and locations on which each image is analyzed, this would already greatly reduce the face-detection execution time. Additionally, the face detector could be applied to only a part of the frames (and, for example, a tracking algorithm could be used for the frames in between), since variation in the location of the face in subsequent frames is expected to be limited. Furthermore, a landmark-detection algorithm could be trained that reaches its final landmark estimation in fewer iterations, uses fewer regression trees per iteration or reduces the depth of each regression tree in order to speed up the detection process.




\ifCLASSOPTIONcaptionsoff
  \newpage
\fi



\bibliographystyle{IEEEtran}

\balance
\bibliography{references}

%


\vfill


\end{document}